\documentclass[letterpaper, 10 pt, conference]{ieeeconf}  

\usepackage{amsmath,graphicx}
\usepackage{multirow}

\usepackage{color}

\IEEEoverridecommandlockouts                              

\title{\LARGE \bf
Calibration-free BEV Representation for Infrastructure Perception
}

\author{Siqi Fan$^{1}$  Zhe Wang$^{1}$  Xiaoliang Huo$^{1,2}$  Yan Wang$^{*1}$ Jingjing Liu$^{1}$
\thanks{$^{*}$ Corresponding author: {\tt\small wangyan@air.tsinghua.edu.cn};}
\thanks{$^{1}$ Institute for AI Industry Research, Tsinghua University;}
\thanks{$^{2}$ Beihang University;}
}

\begin{document}

\maketitle
\thispagestyle{empty}
\pagestyle{empty}

\begin{abstract}

Effective BEV object detection on infrastructure can greatly improve traffic scenes understanding and vehicle-to-infrastructure (V2I) cooperative perception. However, cameras installed on infrastructure have various postures, and previous BEV detection methods rely on accurate calibration, which is difficult for practical applications due to inevitable natural factors (e.g., wind and snow). In this paper, we propose a \textbf{Calibration-free BEV Representation (CBR)} network, which achieves 3D detection based on BEV representation without calibration parameters and additional depth supervision. Specifically, we utilize two multi-layer perceptrons for decoupling the features from perspective view to front view and bird-eye view under boxes-induced foreground supervision. Then, a cross-view feature fusion module matches features from orthogonal views according to similarity and conducts BEV feature enhancement with front view features. Experimental results on DAIR-V2X demonstrate that CBR achieves acceptable performance without any camera parameters and is naturally not affected by calibration noises. We hope CBR can serve as a baseline for future research addressing practical challenges of infrastructure perception.

\end{abstract}

\section{Introduction}
\label{sec:intro}

\begin{figure}
\footnotesize
\begin{minipage}[b]{1.0\linewidth}
  \centering
  \centerline{\includegraphics[scale=0.5]{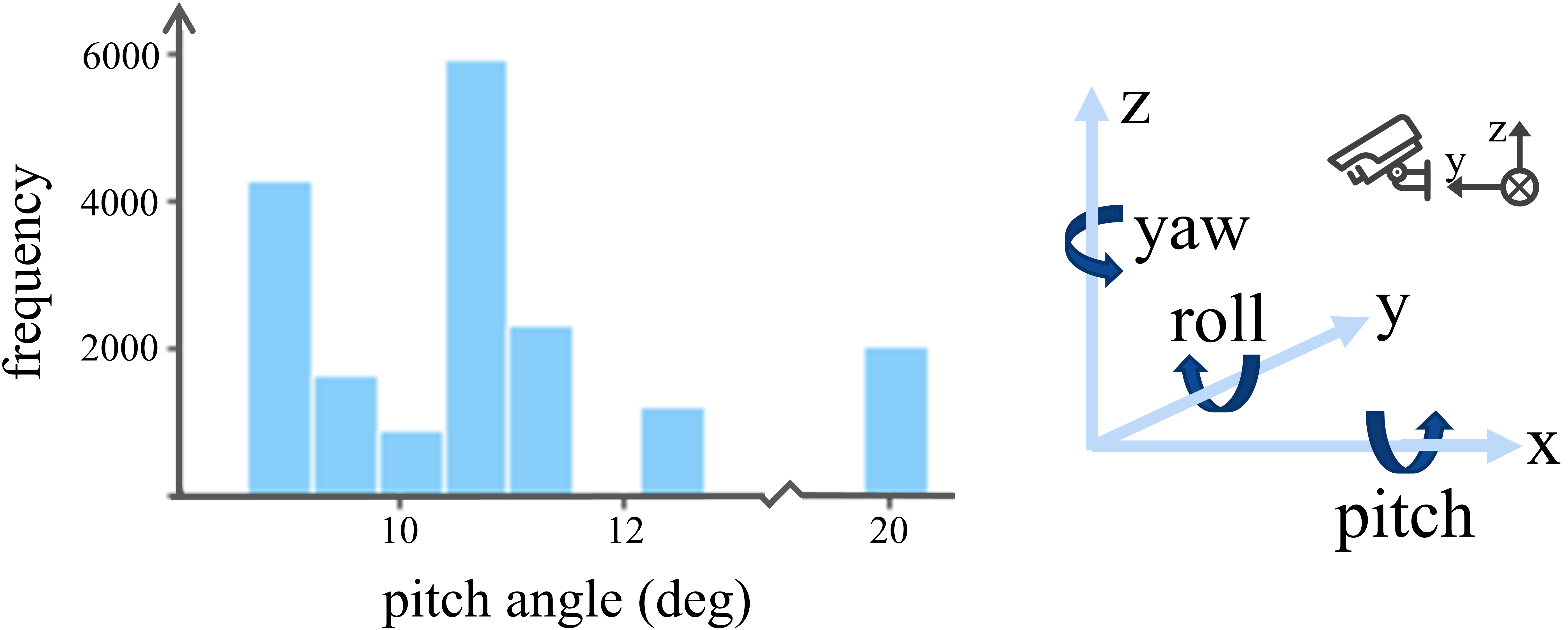}}
  \centerline{(a) The distribution of pitch angles in DAIR-V2X dataset.}\medskip
\end{minipage}
 \vspace{10pt}
\begin{minipage}[b]{1.0\linewidth}
  \centering
  \centerline{\includegraphics[scale=0.45]{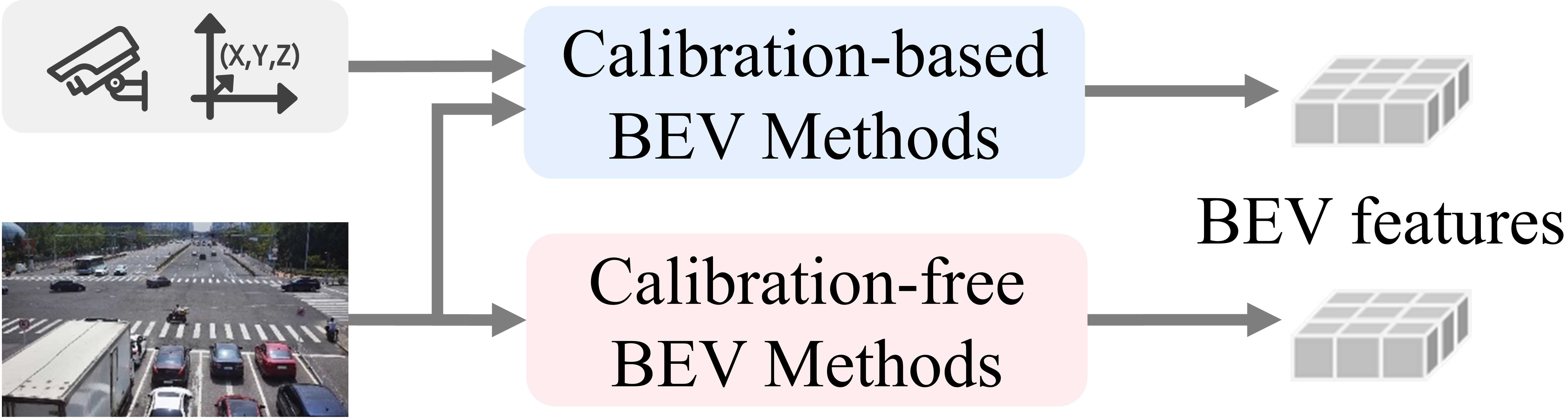}}
  \centerline{(b) Calibration-based and calibration-free BEV methods.}\medskip
\end{minipage}

\caption{Cameras are installed in various postures, especially various pitch angles, on infrastructures (data is from DAIR-V2X dataset \cite{yu2022dair}). Compared with previous calibration-based methods, our approach is calibration-free, which is naturally not constrained by calibration accuracy. }
\label{fig:fig1}
\end{figure}

3D object detection is one of the key enabling technologies for environment perception. Compared with LiDAR-based methods, vision-based methods are cost-effective and easy to implement. However, it is an ill-posed problem to recover 3D information from 2D image.  Existing methods can be grouped into three categories according to when the 2D information is lifted to 3D, including data lifting-based methods, feature lifting-based methods, and result lifting-based methods \cite{ma20223d}. Most of them are not designed specifically for infrastructure side. Different from typical applications, such as vehicle-side environment perception, object detection on infrastructure side has two main challenges: \textit{1) computing resource is limited, 2) cameras are installed in various postures (Figure~\ref{fig:fig1}.a), and accurate calibration parameters are hard to obtain or dynamically correct due to natural factors, like wind and snow}.

Among the aforementioned categories, 2D images are directly transformed into pseudo 3D data (e.g., point cloud) and processed via LiDAR-based pipeline in data lifting-based methods \cite{ma2019accurate, wang2019pseudo, weng2019monocular}, which are computational expensive for infrastructure. Result lifting-based methods \cite{zhang2022monodetr, monoflex, chen2020monopair, liu2020smoke,chen20153d} recover 3D information, including 3D locations and dimensions, based on 2D perspective view features and fully leverage the advantages of 2D detection pipelines. Although they can basically address the aforementioned challenges, the features in perspective view hinder the further development of V2I cooperative perception researches, which only enables result-level fusion. Considering feature-level fusion, feature lifting-based methods \cite{bevdet, bevdepth, bevformer, petr, imvoxelnet, reading2021categorical} first transform 2D image feature into 3D voxel feature and collapse them to generate BEV features. BEV features from different agents, time series, and modalities can be fused in a physics-interpretable manner \cite{bevsurvey}, and 3D detection based on BEV representations has attracted immense attention in recent years. However, these methods rely on accurate camera calibration (i.e. intrinsic and extrinsic parameters) and/or additional depth supervision to assist cross-view feature projection, which are not suitable for infrastructure-side perception because of unavoidable calibration noise, and their performance would be significantly degraded if using noisy parameters. PYVA \cite{pyva} generated BEV representations via multi-layer perceptrons (MLPs) without camera parameters for road scene layout estimation on vehicle-side, but the performance is far from satisfaction when migrated to detection task on infrastructure side.

To address the practical challenges of infrastructure-side perception, we propose a Calibration-free BEV Representation (CBR) network, which is naturally not constrained by calibration accuracy (Figure~\ref{fig:fig1}.b). Specifically, we utilize light image backbone ResNet-18 to extract perspective view feature. In the face of various camera postures, two MLPs are used for view decoupling from perspective view to front and bird-eye view. The view transformation is supervised by boxes-induced foreground segmentation labels generated from 3D bounding box labels, without additional labeling cost. To compensate information loss of BEV features along height dimension (z-axis), a cross-view feature fusion module is proposed for BEV feature enhancement using front view features. Assuming that same object should have similar features under different views, features from  orthogonal views are fused according to similarity distribution. 

Our contributions are summarized as follows:
\begin{itemize}
  \setlength{\itemsep}{0pt}
  \setlength{\parsep}{0pt}
  \setlength{\parskip}{0pt}
  \item We point out the practical challenges of infrastructure-side perception, and propose the Calibration-free BEV Representation network, CBR, to address various install postures and calibration noise. 
  \item The perspective view features are decoupled to front view and bird-eye view via MLPs without any calibration parameters, and orthogonal feature fusion is similarity-based without additional depth supervision.
  \item Experimental results demonstrate that CBR achieves acceptable detection performance based on BEV representation on large-scale real-world dataset DAIR-V2X and can output BEV foreground segmentation predictions at the meantime.
\end{itemize}

\section{Related Work}
In this section, we briefly review two related topics, BEV object detection and BEV representation generation.
\subsection{Image-based BEV Object Detection}
BEV object detection methods have attracted more attention recently, and have made great progress in performance. However, most of them are designed for typical vehicle-side perception and not suitable for infrastructure side. 
Depth-based methods \cite{bevdet, bevdepth, caddn} infer depth to recover 3D information along y-axis of BEV coordinate system, but the depth in image view on infrastructure is the compound information along y-axis and z-axis due to the pitch angle, which cannot be directly utilized for BEV detection. Projection-based methods \cite{imvoxelnet, OFT} are not affected by camera postures, since the features are projected to 3D according to calibration parameters before fed into detection heads, nevertheless, their performance highly relies on calibration accuracy. Transformer-based methods \cite{saha2022translating, petr, bevformer, bevformer2} achieve better performance with higher computational cost, and calibration parameters may also be needed for attention guidance.
 To address the practical challenges of infrastructure-side perception and get rid of the dependence on calibration accuracy, we propose CBR to achieve 3D perception in a calibration-free manner. 

\subsection{BEV Representation Generation}
With the advantage of succinct and physics-interpretable, BEV representations are deployed in more and more downstream real-world applications, especially for traffic scenes. Despite the aforementioned approaches adopted in detection tasks, how to generate BEV representation from image is also well-studied in segmentation researches, consisting of geometry-based (homograph or depth) and learning-based (MLP or transformer) approaches \cite{bevsurvey}. Homograph-based methods \cite{ipm} realize view projection relying on physical mapping under horizontal plane constraint. Depth-based methods \cite{OFT, caddn, lss, bevdepth, bevdet} explicitly leverage depth distribution to lift 2D features to 3D space (e.g. voxel and points cloud), and depth supervision is an essential cue to them. Learning-based approaches ignore the geometric priors from calibrations. MLP-based methods \cite{vpn, pon, pyva} model the transformation via the global mapping capability of MLP. On account of strong modeling ability, transformer-based methods \cite{petr, bevformer, bevformer2} are further developed recently. It would be a considerable option for devices with sufficient computing resources. In this paper, MLP is used for view decoupling, and similarity-based cross-view fusion is proposed taking inspiration from depth-based methods.

\section{CBR Framework}

\begin{figure*}
  \centering
  \includegraphics[scale=0.47]{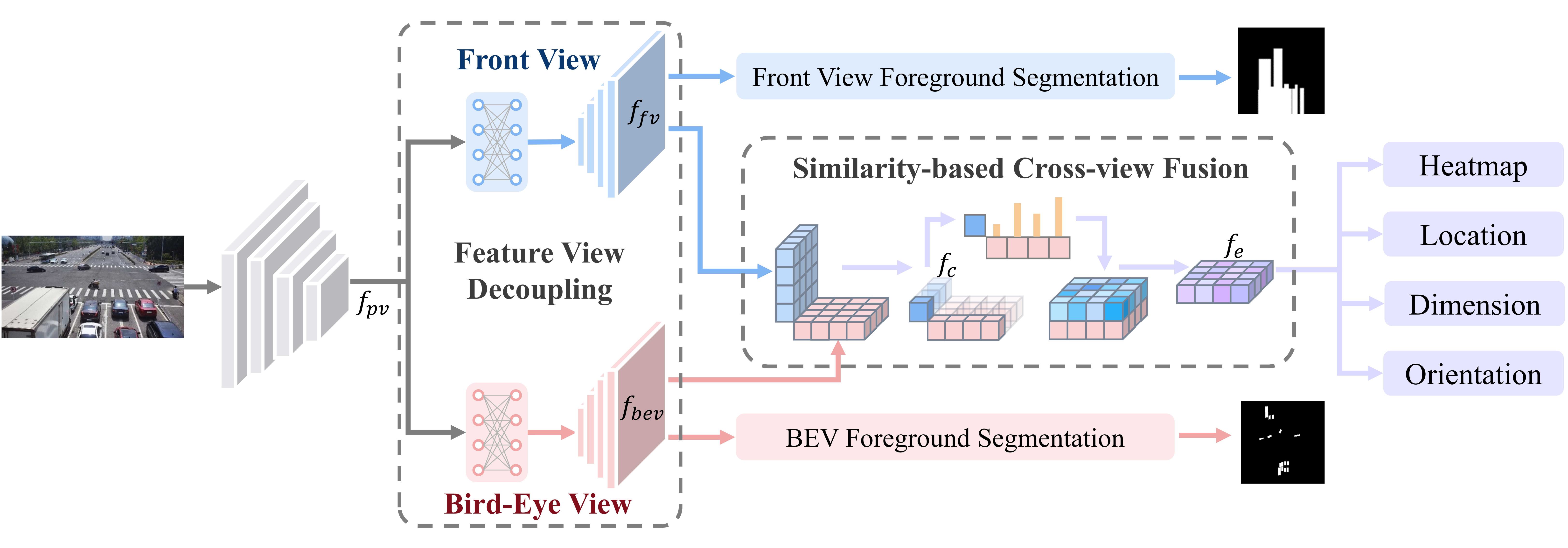}
  \caption{Framework of the Calibration-free BEV Representation (CBR) network.}
  \label{fig:CBR}
\end{figure*}

\label{sec:method}
In this section, we describe the proposed calibration-free BEV representation network, which mainly consists of  feature view decoupling module and  similarity-based cross-view fusion module.

\subsection{Overall Architecture}
Addressing practical challenges, CBR achieves feature view standardization via decoupled feature reconstruction (Figure~\ref{fig:CBR}). Images captured from infrastructure are fed into an image backbone to extract perspective view features. With the consideration of limited computing resources, ResNet-18 is employed. The feature maps are further processed by a convolution operation with a filter in the size of $3 \times 3$ and a mean pooling operation to save the computational cost of the view decoupling. Image scale is gradually decreased from $H \times W$ to $\frac{H}{64} \times \frac{W}{64}$, while the channel size is increased to $1024$. Taking the advantage of global receptive field, the perspective view feature $f_{pv}$ is spatially decoupled to two orthogonal views via FVD (feature view decoupling) module. Next, a SCF (similarity-based cross-view fusion) module is used to match the features from different views and generate enhanced BEV features $f_e$ leveraging front view features $f_{fv}$ and bird-eye view features $f_{bev}$. Finally, $f_e$ is fed to $4$ detection heads for classification and regression tasks. Each detection head is composed of a 
basic convolution block, including convolution, batch normalization, and RELU.

\subsection{Feature View Decoupling}
In real-world scenes, cameras installed on infrastructure side usually have various postures, including x-y-z location and pitch-yaw-roll orientation. Compared with location, orientation, especially pitch, will directly affect the perspective view features. To generate unified representations despite of various orientation, we propose the FVD module for feature view decoupling. Since the features of different views are not spatially aligned, MLP can better facilitate the view decoupling compared with convolution operation. The MLP structure consisted of two fully connected layers is deployed following the practice of previous works \cite{pan2020cross, pyva}. The decoupled features are fed to four consecutive decoder layers, and we utilize nearest interpolation for upsampling from $\frac{H}{64} \times \frac{W}{64}$ to $\frac{H}{4} \times \frac{W}{4}$. After that, the front view features $f_{fv}$ and BEV features $f_{bev}$ are obtained:

\begin{equation}
  f_{fv} = \Phi_{fv}(\mathit{MLP}(f_{pv}))
\end{equation}
\begin{equation}
  f_{bev} = \Phi_{bev}(\mathit{MLP}(f_{pv}))
\end{equation}
where `$\Phi(\cdot)$' denotes the feature decoding operation, and $f_{pv}$ is the extracted features in perspective view. 

To guide the view decoupling without using calibration parameters, $f_{fv}$ and $f_{bev}$ are further input to the corresponding foreground segmentation heads (composed of a basic convolution block), respectively. The segmentation prediction is under the boxes-induced foreground supervision, which is generated by projecting the 3D bounding boxes to front/bird-eye view, without additional labeling cost. The benefits of the foreground segmentation supervision are two-fold. On the one hand, the pixel-level supervision can effectively guide the view transformation and encourage the module focus on foreground objects (e.g., cars). On the other hand, the BEV foreground segmentation predictions are output as by-product, which indicates the dynamic foreground layout of the traffic scenes.

\subsection{Similarity-based Cross-view Fusion}
BEV features can effectively represent the foreground layout in bird-eye view. However, 3D detection performance based on that will naturally be influenced by the information loss along z-axis, especially when the view projection is not accurately guided with calibration parameters. Therefore, it is necessary to enhance the BEV representations with features in the front view, and the main difficulty is matching the corresponding features across orthogonal views. 

\begin{figure}
\footnotesize
\begin{minipage}[b]{1.0\linewidth}
  \centering
  \centerline{\includegraphics[scale=0.4]{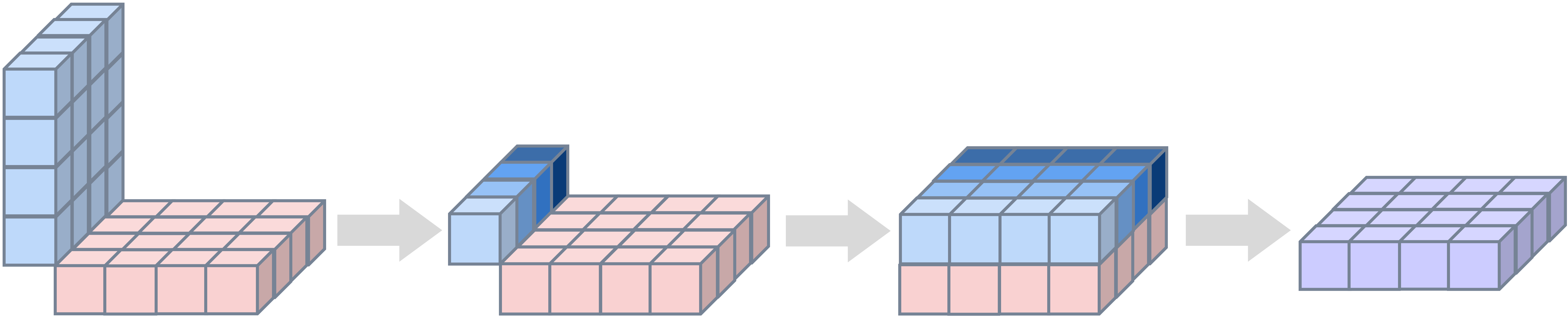}}
  \centerline{(a) Condense-Push fusion (CPF).}\medskip
\end{minipage}
 \vspace{1pt}
\begin{minipage}[b]{1.0\linewidth}
  \centering
  \centerline{\includegraphics[scale=0.4]{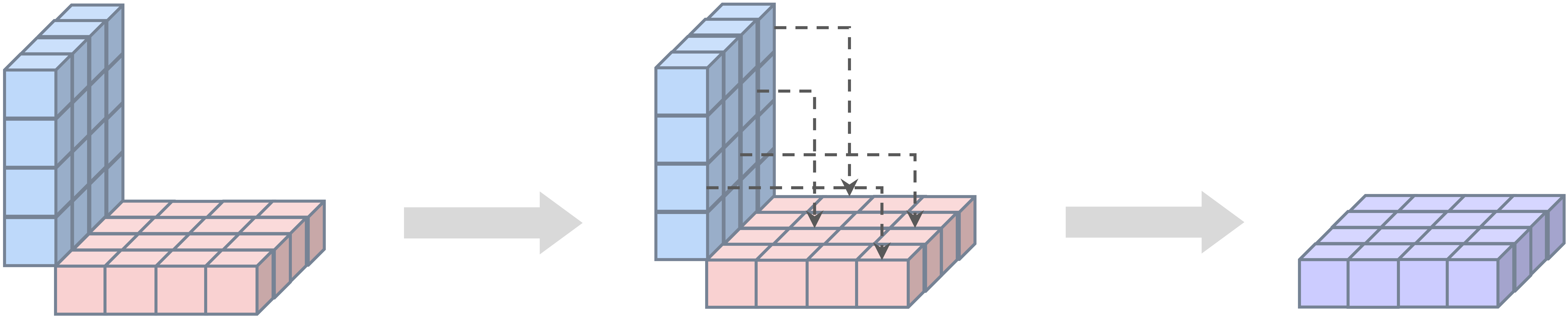}}
  \centerline{(b) Similarity-based global fusion (SGF).}\medskip
\end{minipage}

\caption{Illustration of condense-push fusion (CPF) and similarity-based global fusion (SGF).}
\label{fig:fusion}
\end{figure}

There are two heuristic options, as shown in Figure~\ref{fig:fusion}. Assuming that the feature of the same object in different views should be similar, SGF (similarity-based global fusion) can match features globally according to similarity, but it is computational expensive. To reduce the searching space of feature matching between two views, CPF (condense-push fusion) first condenses $f_{fv}$ along z-axis and then pushes the obtained $f_c$ along 
y-axis making use of geometric constrains. 

To embrace the advantages of both CPF and SGF, we design SCF (similarity-based cross-view fusion) module, which matches the features based on similarity with the geometric constrains (Figure~\ref{fig:CBR}). Specifically, we only take the similarity among the features with the same x-axis value into consideration. To reduce computational cost, we utilize the condensed feature $f_c= \mathit{Avg}(f_{fv})$ for feature fusion, where `$\mathit{Avg}(\cdot)$' denotes mean pooling operation along z-axis. The similarity $s_{ij}$ is measured by the inner-product:
\begin{equation}
  s_{ij} = \langle f_{c_i}, f_{bev_{ij}}  \rangle
\end{equation}
where `$\langle \cdot, \cdot \rangle$' denotes inner-product operation, and $i$ is the index of x-axis, $j$ is the index along 
y-axis. The calculated similarity is used as fusion weights to enhance BEV feature $f_{bev}$ with condensed front view feature $f_c$:
\begin{equation}
  f_e = \mathit{Conv}(\mathit{Concat}(f_{bev}, s \cdot f_c))
\end{equation}
where `$\mathit{Conv}(\cdot)$' and `$\mathit{Concat}(\cdot)$' denote convolution and concatenation operations.

\begin{figure*}
  \centering
  \includegraphics[scale=0.43]{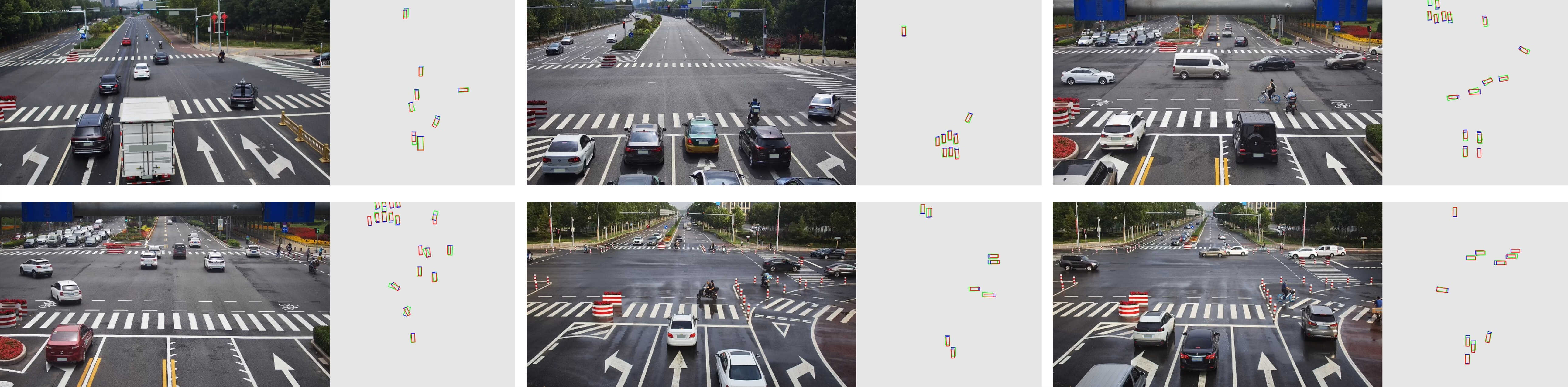}
  \caption{Visualization examples. Red: groundtruth. Green: predictions of CBR. Blue line indicates the head of vehicle.}
  \label{fig:demo}
\end{figure*}

\begin{table*}
  \centering
  \caption{Quantitative evaluation on DAIR-V2X Dataset with calibration noise on rotation angles. The performance is significantly degraded with noisy calibration parameters, while our approach is not influenced. All scores are in $\%$.}
  \begin{tabular}{c|c|ccc|ccc|ccc|ccc} 
  \hline
    \multirow{2}*{\textbf{Methods}}  & \multirow{2}*{\textbf{Calib. Noise (deg)}}  & \multicolumn{3}{c|}{$AP_{3D|R40}$ (IoU=0.5)}  & \multicolumn{3}{c|}{$AP_{3D|R40}$ (IoU=0.7)}  & \multicolumn{3}{c|}{$AP_{BEV|R40}$ (IoU=0.5)}  & \multicolumn{3}{c}{$AP_{BEV|R40}$ (IoU=0.7)}  \\
       &   & \textbf{easy}   & \textbf{mod.}   & \textbf{hard}   & \textbf{easy}   & \textbf{mod.}   & \textbf{hard}   & \textbf{easy}   & \textbf{mod.}   & \textbf{hard}   & \textbf{easy}   & \textbf{mod.}   & \textbf{hard} \\ \hline

    \multirow{7}*{ImVoxelNet \cite{imvoxelnet}} & / & 47.6 & 29.2 & 27.1 & 27.1 & 16.2 & 14.8 & 51.9 & 32.7 & 30.4 & 35.4 & 20.5 & 20.1 \\
    
      & 0.1 & 44.5 & 26.5 & 26.2 & 24.2 & 13.9 & 12.5 & 50.9 & 32.0 & 29.9 & 32.1 & 19.0 & 17.5 \\

      & 0.2 & 38.6 & 23.1 & 22.6 & 21.0 & 11.5 & 11.2 & 45.1 & 26.8 & 26.4 & 27.6 & 16.4 & 15.0 \\

      & 0.5 & 29.3 & 16.9 & 15.4 & 12.9 & 6.8 & 6.5 & 35.0 & 20.1 & 19.7 & 19.1 & 10.9 & 9.9 \\

      & 1.0 & 19.7 & 11.4 & 10.2 & 5.0 & 2.4 & 2.4 & 25.5 & 14.7 & 14.3 & 9.6 & 5.3 & 4.7 \\

      & 2.0 & 8.2 & 4.4 & 4.3 & 1.0 & 0.5 & 0.5 & 13.6 & 7.2 & 7.0 & 2.2 & 1.0 & 1.0 \\

      & 5.0 & 0.6 & 0.3 & 0.3 & 0.0 & 0.0 & 0.0 & 1.4 & 0.7 & 0.7 & 0.1 & 0.1 & 0.1 \\ \hline

    PYVA-det & calibration-free & 12.6 & 7.3 & 7.1 & 0.9 & 0.6 & 0.5 & 23.3 & 14.0 & 13.6 & 5.5 & 2.9 & 2.9 \\ \hline

    CBR (Ours) & calibration-free & 24.7 & 15.7 & 14.7 & 1.3 & 0.8 & 0.8 & 40.0 & 24.9 & 24.5 & 4.9 & 3.2 & 3.2 \\ \hline
    
  \end{tabular}
  \label{tab:quant}
\end{table*}

It can be presumed that the feature similarity distribution across orthogonal views along y-axis is implicit depth distribution, since the closer to the real depth, the more similar the features across views are. SCF bridges the cross-view features without additional depth supervision. Moreover, the similarity-based fusion indirectly facilitates the spatial-wise alignment across different views, as the corresponding features are encouraged to be in the same $x$ column.

\section{Experiments}
This section describes experiments on real-world infrastructure detection dataset. We compare our model with other typical BEV detection methods in noisy calibration setting, and provide detailed ablation study and error analysis. Further evaluation on BEV foreground segmentation also validates the scene layout understanding capacity of CBR.

\subsection{Experimental Setting}
\textbf{Datasets}
We evaluate the proposed CBR model on the large-scale real-world cooperative perception dataset DAIR-V2X \cite{yu2022dair}. It provides 12,424 images captured from diverse infrastructure-side cameras with 3D annotations, which comprises 8800 images for training and 3624 images for validation. We follow the official split scheme and report experimental results on validation set. All of the objects inside the camera view are labeled, and the perception range of our method is set as $90m \times 90m \times 5m$. The input images are resized to a fixed size of $1024 \times 1024$.

\textbf{Foreground Segmentation Label Generation}
To generate foreground segmentation labels in orthogonal views, each bounding box of labeled objects in perception range is projected to bird-eye and front view. The generated pixel-level groundtruth is with the size of $256 \times 256$.

\textbf{Calibration Noise}
To simulate the natural calibration noise in practical environments, we introduce several levels of Gaussian noise to rotation angles
\begin{equation}
  \theta_n = x_n * n_{range}
\end{equation}
where $x_n \sim N(\mu, \sigma^2)$, $\mu = 0, \sigma = \frac{1}{3}$, and $n_{range} \in \{0.1, 0.2, 0.5, 1.0, 2.0, 5.0\}$ denotes the noise level in degree.

\textbf{Baselines}
We compare CBR with both calibration-based and calibration-free BEV methods. ImVoxelNet \cite{imvoxelnet} is a typical projection-based detection method, which projects features from perspective view to BEV with the guidance of calibration parameters. PYVA \cite{pyva} is originally proposed for calibration-free segmentation. We develop PYVA-det based on that by adopting additional detection heads.

\textbf{Implementation Details}
The image backbone, ResNet 18, is initialized with pre-trained weights provided by PyTorch. The initial learning rate is set to $2 \times 10^{-4}$. We use adam optimizer for training with batch size of 6, and the network is trained until convergence (200 epochs). Random flipping and random color jitter are applied while training. Experiments are implemented in PyTorch on a server with NVIDIA A30.

\subsection{Main Results}

Table~\ref{tab:quant} reports the comparison results between CBR and baselines with calibration noise. The performance with the IoU threshold of $0.5$ is regarded as major concern. CBR achieves $15.7\%$ and $24.9\%$ mAP on 3D detection and BEV detection tasks for moderate difficulty. Some visualization examples are shown in Figure~\ref{fig:demo}. 

\begin{figure}[ht]
  \centering
  \includegraphics[scale=0.45]{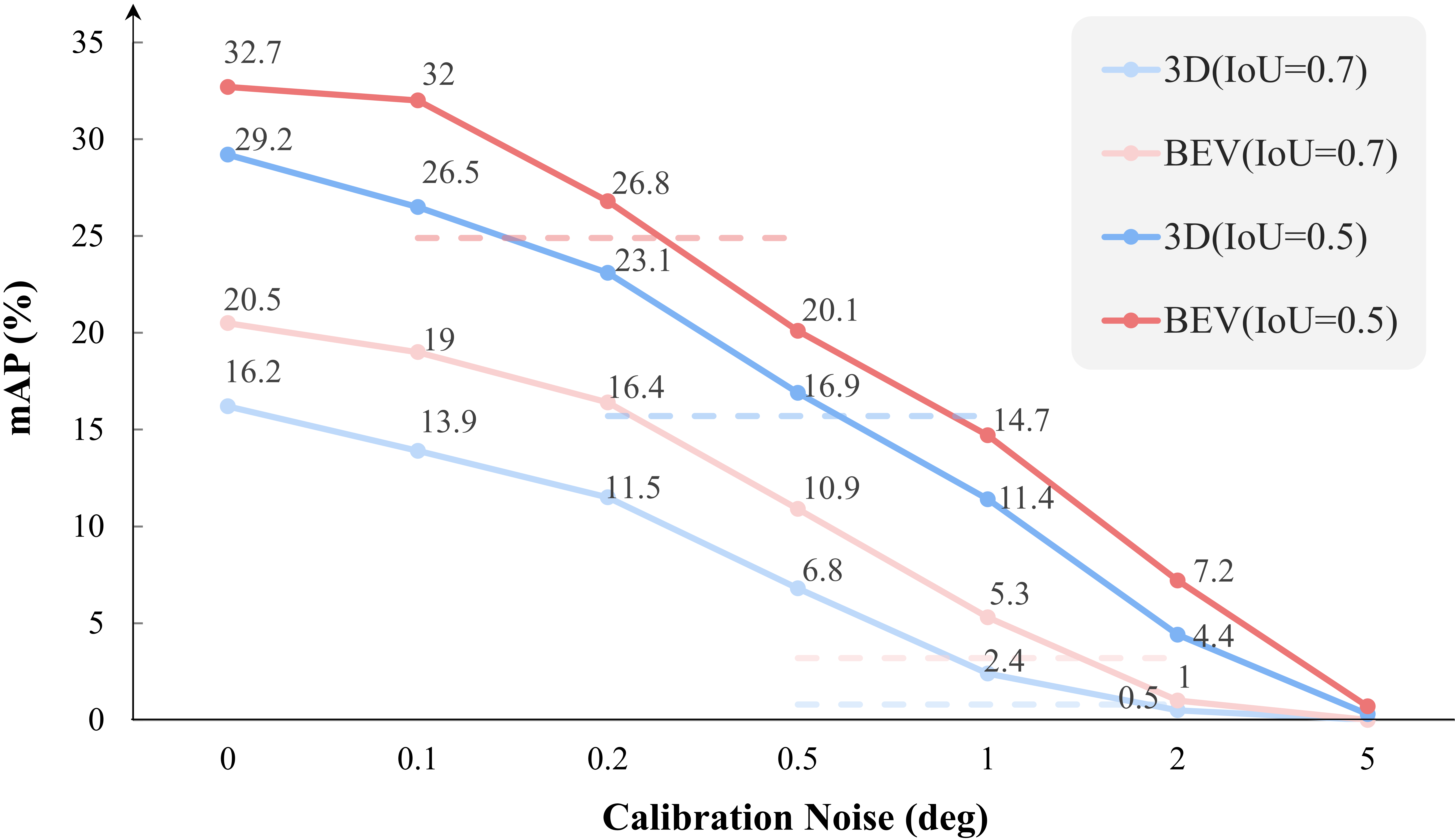}
  \caption{Comparison between calibration-based method \cite{imvoxelnet} (solid line) and our CBR (dotted line) with calibration noise under moderate difficulty. CBR is naturally not affected by noisy rotation angles.}
  \label{fig:com}
\end{figure}

It can be seen that ImVoxelNet \cite{imvoxelnet} performs better leveraging accurate calibration parameters. However, the performance is significantly degraded while noise raises, almost cut in half if small rotation noise (within 0.5 degree) is introduced randomly. Different from that, calibration-free methods are naturally not affected by noisy calibration parameters, and show the superiority in noisy cases, visually shown in Figure~\ref{fig:com}. Our CBR has performance advantage on BEV detection, when $n_{range}$ is greater than $0.2$ degree. The watershed on 3D detection task is $0.5$ degree. We also report experimental results with a more strict IoU threshold, $IoU=0.7$, and performance trend in noisy case is in line with that of $IoU=0.5$. Since calibration noise on infrastructure side is unavoidable due to the complex natural factors (e.g. wind and snow), calibration-free methods are more robust under various scenes. 

Although PYVA-det is not limited by calibration accuracy either, the performances on 3D/BEV tasks are far from satisfaction. Our approach achieves better accuracy-robustness balance for infrastructure perception.

\subsection{Ablation Study}

We conduct the following experiments to study the impact of different cross-view feature fusion methods. The performance comparisons are summarized in Table~\ref{tab:ablation}, and the baseline is vanilla BEV representation, which is directly obtained via feature view decoupling without cross-view feature fusion. Benefiting from similarity-based fusion, SGF effectively leads to an improvement of $0.9\%$ on BEV detection, but the performance growth on 3D detection is limited. CPF performs better than SGF leveraging the front view features with geometric constrains, and the performance on 3D task is lifted to $13.6\%$. Embracing advantages of geometry and similarity, SCF further boosts the performances on both tasks, and achieves $15.7\%$ and $24.9\%$ respectively. The performance advantage of SCF is obvious compared with SGF and CPF.

\begin{table}[ht]
  \centering
  \caption{Ablation study on cross-view feature fusion.}
  \begin{tabular}{c|ccc|ccc} 
  \hline
    \multirow{2}*{\textbf{Ablated}} & \multicolumn{3}{c|}{$AP_{3D|R40}$ (IoU 0.5)} & \multicolumn{3}{c}{$AP_{BEV|R40}$ (IoU 0.5)} \\ 

     & \textbf{easy} & \textbf{mod.} & \textbf{hard} & \textbf{easy} & \textbf{mod.} & \textbf{hard} \\ \hline

     Vanilla-BEV & 20.2 & 12.6 & 11.7 & 36.8 & 22.2 & 20.7 \\
     SGF & 21.8 & 13.0 & 12.8 & 37.3 & 23.1 & 22.7 \\
     CPF & 21.8 & 13.6 & 13.3 & 38.9 & 23.5 & 23.1 \\
     SCF & 24.7 & 15.7 & 14.7 & 40.0 & 24.9 & 24.5 \\ \hline
    
  \end{tabular}
  \label{tab:ablation}
\end{table}

\begin{figure}
  \centering
  \includegraphics[scale=0.45]{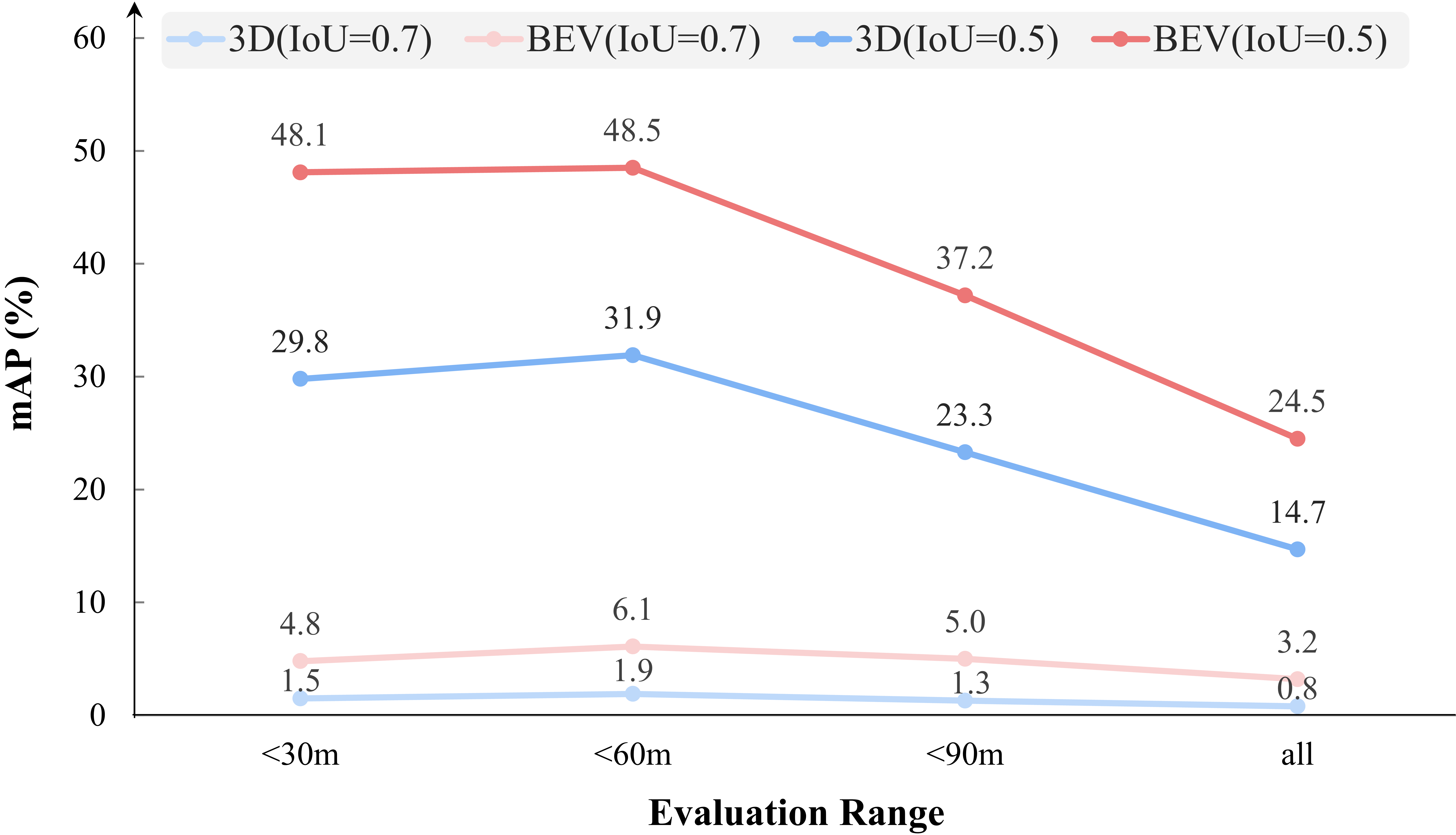}
  \caption{Error analysis: evaluation with distance.}
  \label{fig:dis}
\end{figure}

\subsection{Error Analysis}

To better understand CBR, we further conduct several groups of experiments for error analysis, and the limitations of calibration-free BEV representation are discussed.

\textbf{Evaluation with distance.}
The following experiments are conducted to analysis the performance under different perception range (Figure~\ref{fig:dis}). The infrastructure-side perception region is split to four parts with three thresholds of 30m, 60m, and 90m. It can be seen that the performance degradation is inevitable along with distance increasement. The performance is almost doubled if we only take the objects within the range of 60m into consideration. Note that the performance within 60m is slightly better than that within 30m. We think this increment is caused by the amount of information of vehicles in different views. Specifically, the object within 30m captured from infrastructure is almost in a top view, while it tends to be in side view when extended to 60m. Intuitively, the side view is more informative than top view. 
In addition, the decline is obvious out of the range of 90m, which is the designed representation range of our BEV feature. Therefore, detection capability is limited by the the manually set perception range, and the objects lie out of that are theoretically ignored.

\begin{figure}
\footnotesize
\begin{minipage}[b]{1.0\linewidth}
  \centering
  \centerline{\includegraphics[scale=0.42]{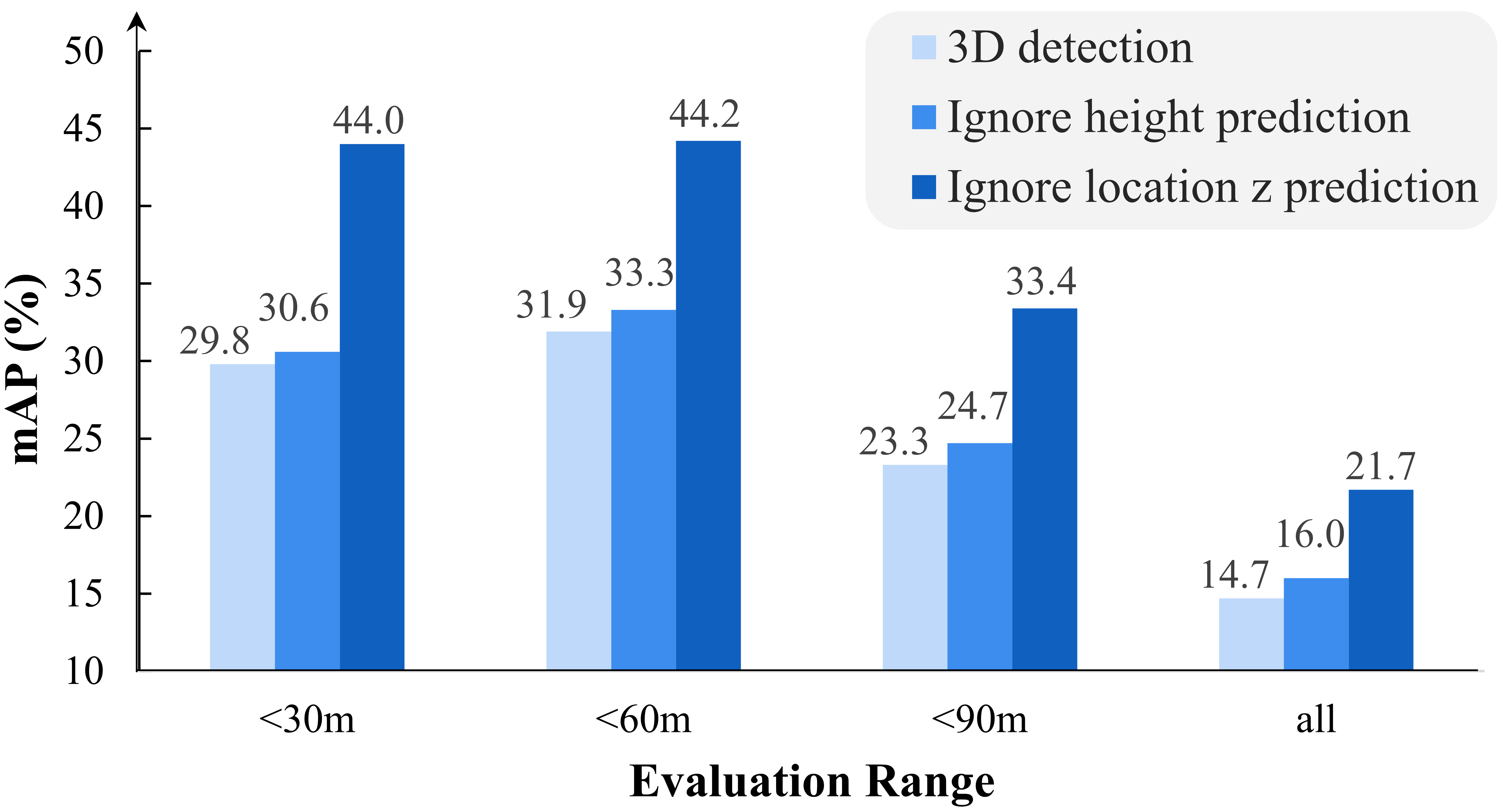}}
  \centerline{(a) 3D detection performance (IoU 0.5).}\medskip
\end{minipage}
 \vspace{1pt}
\begin{minipage}[b]{1.0\linewidth}
  \centering
  \centerline{\includegraphics[scale=0.42]{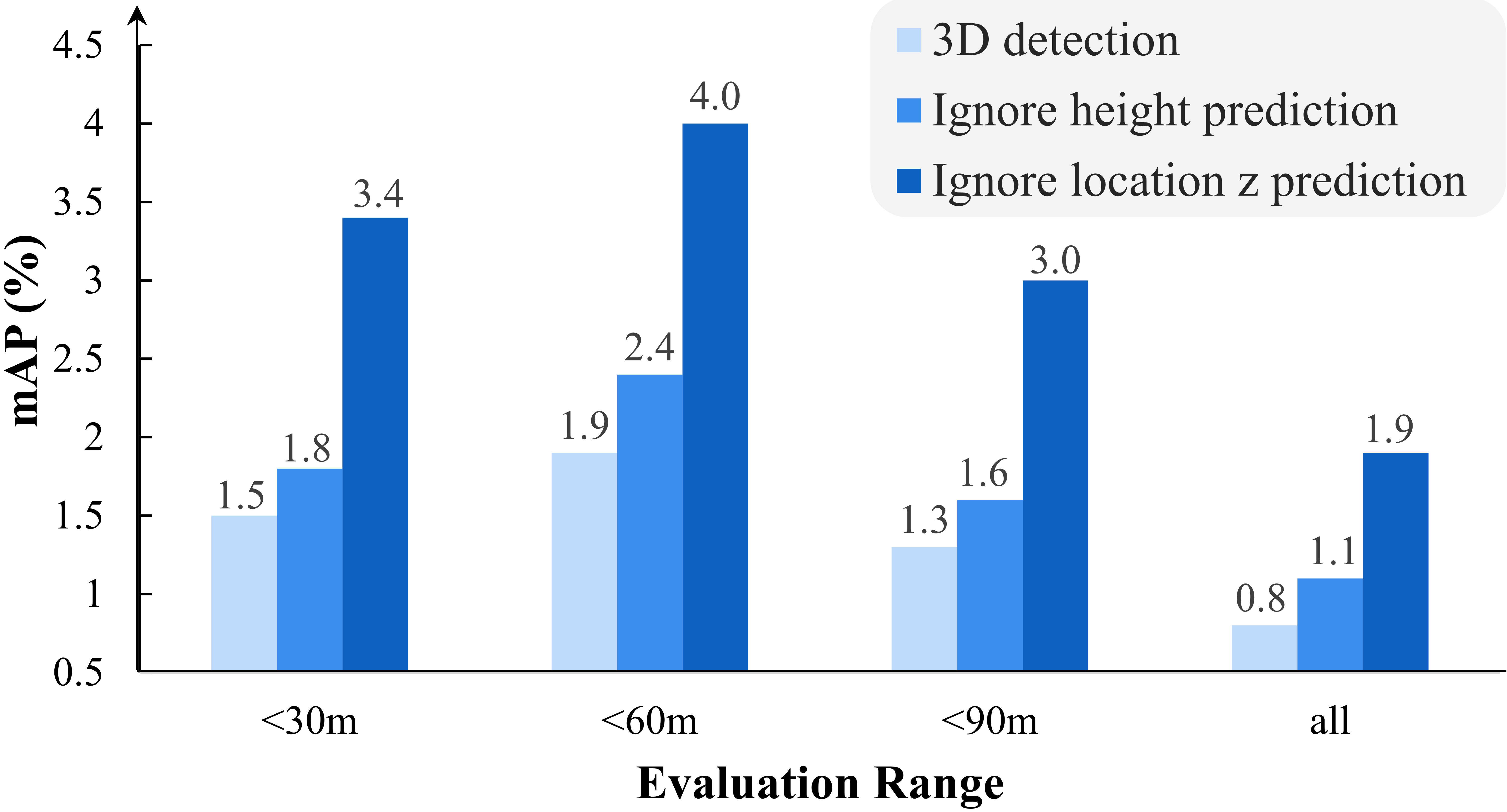}}
  \centerline{(b) 3D detection performance (IoU 0.7).}\medskip
\end{minipage}

\caption{Error analysis: major error source of 3D detection.}
\label{fig:source}
\end{figure}

\textbf{Error sources of 3D detection.}
Compared with BEV detection, performance on 3D detection is worse since the additional prediction along z-axis, including location-z and height. To analysis the major source of error, we evaluated the ablated predictions by ignoring either location-z or height predictions. As shown in Figure~\ref{fig:source}, the score increment is more obvious if ignoring predictions of location-z rather than height predictions, regardless of the IoU threshold (0.5 or 0.7), which indicates location-z prediction is the major error source on 3D detection. It is difficult to estimate the location along z axis for infrastructure perception due to the various installation height of cameras, especially without the reference of calibration parameters.

\subsection{Evaluation on BEV Foreground Segmentation}
We further evaluate CBR on BEV foreground segmentation, which is a by-product of feature view decoupling (Table~\ref{tab:seg}). The performance of PYVA \cite{pyva} is declined when additional detection heads are introduced. Although CBR is slightly worse than PYVA, it is still better than PYVA-det.

\begin{table}[ht]
  \centering
  \caption{Quantitative evaluation on BEV foreground segmentation.}
  \begin{tabular}{c|cc} 
  \hline
    \textbf{Methods} & \textbf{mIoU (\%)} & \textbf{mAP (\%)} \\ \hline
    PYVA \cite{pyva} & 42.3 & 56.0 \\
    PYVA-det & 33.9 & 43.5 \\ \hline
    CBR (Ours) & 40.3 & 50.4 \\ \hline
    
  \end{tabular}
  \label{tab:seg}
\end{table}

\section{Conclusion}
Addressing the practical challenges of various installation postures and calibration noises, we point out the significant performance degradation of calibration-based BEV detection approach under calibration noise, and propose a calibration-free BEV representation for infrastructure perception in this paper. The extracted image features are decoupled to two orthogonal views, and BEV representations are enhanced via similarity-based cross-view fusion. Extensive experiments on real-world dataset demonstrate that CBR achieves a better accuracy-robustness balance. In addition, error analysis are reported, and limitations of the proposed calibration-free BEV representations are further discussed. In future work, the way to utilize partial stable calibration parameters to improve perception performance deserves to be studied, and how to leverage multi-view images for adaptive camera re-calibration is also worth to be further explored.

\newpage
\section*{\textbf{Appendix: more experimental results}}

CBR is proposed for practical infrastructure perception to facilitate the development of V2I cooperative perception. The experiments in Sec.IV are conducted on DAIR-V2X-C to provide an infrastructure-side baseline for VIC3D task, and we further report more experimental results on DAIR-V2X-I to make comparison with more recent approaches.

\textbf{Datasets}
Different from the main dataset DAIR-V2X-C for V2I cooperative perception, DAIR-V2X-I only contains infrastructure-side data. It includes around 10 thousand frames and is divided into train/val/test (50/20/30) subsets. We evaluate CBR on validation set following \cite{bevheight}.

\textbf{Baselines}
We compare our CBR with other SOTA camera-based methods like ImVoxelNet \cite{imvoxelnet}, BEVFormer \cite{bevformer}, BEVDepth \cite{bevdepth}, and BEVHeight \cite{bevheight}. In addition, some LiDAR-based and multimodal-based methods are also reported for reference, including PointPillars \cite{pointpillars}, SECOND \cite{second}, and MVXNet \cite{mvxnet}.

\begin{figure*}
  \centering
  \includegraphics[scale=0.43]{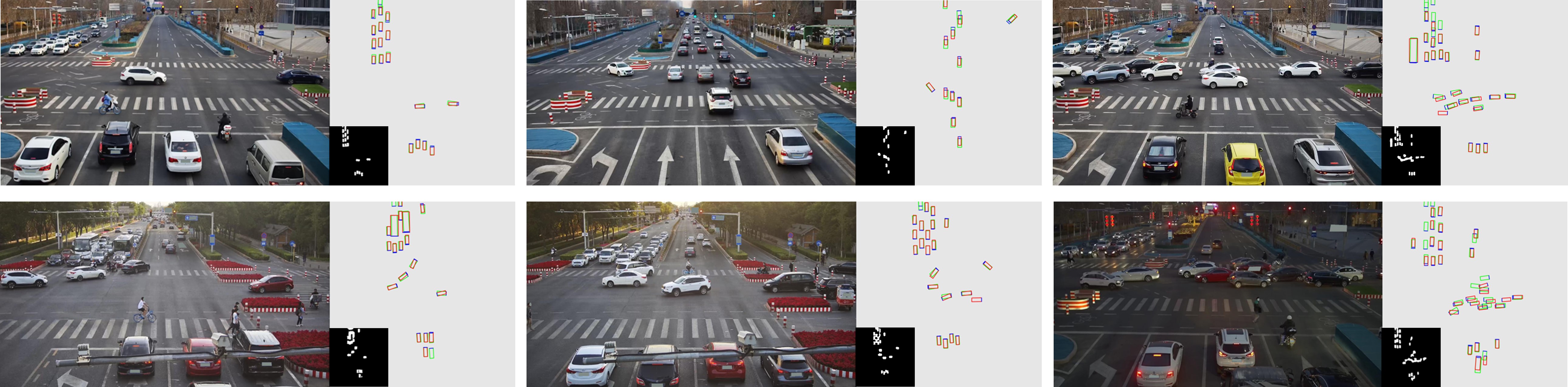}
  \caption{Visualization examples on DAIR-V2X-I. Red: groundtruth. Green: predictions of CBR. Blue line indicates the head of vehicle.}
  \label{fig:demo_i}
\end{figure*}

\textbf{Experimental results on DAIR-V2X-I}
We report both 3D and BEV perception performance under two IoU threshold in Table~\ref{tab:apx1}. CBR achieves $60.1\%$ $AP_{3D|R40}$ and $64.5\%$  $AP_{BEV|R40}$ for moderate difficulty, which demonstrates the effectiveness of our method in practical application. Besides, it also achieves $57.9\%$ mIoU and $60.2\%$ mAP on BEV foreground segmentation. Some visualization examples are shown in Figure~\ref{fig:demo_i}.

\begin{table}[ht]
  \centering
  \caption{Experimental results on DAIR-V2X-I.}
  \begin{tabular}{c|ccc|ccc} 
  \hline
    \multirow{2}*{\textbf{Task}} & \multicolumn{3}{c|}{\textbf{IoU=0.5}} & \multicolumn{3}{c}{\textbf{IoU=0.7}} \\ 

     & \textbf{easy} & \textbf{mod.} & \textbf{hard} & \textbf{easy} & \textbf{mod.} & \textbf{hard} \\ \hline

     $AP_{3D|R40}$  & 72.0 & 60.1 & 60.1 & 38.5 & 31.6 & 31.7 \\
     $AP_{BEV|R40}$ & 78.7 & 64.5 & 64.6 & 56.5 & 46.1 & 46.2 \\ \hline
    
  \end{tabular}
  \label{tab:apx1}
\end{table}

\textbf{Comparison on DAIR-V2X-I benchmark}
We compare CBR with recent methods on the original benchmark, as shown in Table~\ref{tab:apx2}. The experimental results of others are from \cite{bevheight}, and all of them are calibration-based. It can be seen that CBR outperforms five of them without the limitation of accurate extrinsic parameters. 

\begin{table}[ht]
  \centering
  \caption{Comparison on DAIR-V2X-I benchmark. The results of others are from \cite{bevheight}. `L' and `C' denote LiDAR and camera. }
  \begin{tabular}{c|c|ccc} 
  \hline
    \multirow{2}*{\textbf{Method}} & \multirow{2}*{\textbf{Modality}} & \multicolumn{3}{c}{$AP_{3D|R40}$ (IoU=0.5)} \\ 

                                   &                   & \textbf{easy} & \textbf{mod.} & \textbf{hard} \\ \hline

     PointPillars \cite{pointpillars} & L              & 63.1          & 54.0          & 54.0 \\
     SECOND \cite{second}             & L              & 71.5          & 54.0          & 54.0 \\ 
     MVXNet \cite{mvxnet}             & LC             & 71.0          & 53.7          & 53.8 \\
     ImVoxelNet \cite{imvoxelnet}     & C              & 44.8          & 37.6          & 37.6 \\
     BEVFormer \cite{bevformer}       & C              & 61.4          & 50.7          & 50.7 \\
     BEVDepth \cite{bevdepth}         & C              & 75.5          & 63.6          & 63.7 \\
     BEVHeight \cite{bevheight}       & C              & 77.8          & 65.8          & 65.9 \\ \hline
     CBR (Ours)                       & C              & 72.0          & 60.1          & 60.1 \\ \hline

  \end{tabular}
  \label{tab:apx2}
\end{table}

\newpage

\bibliographystyle{IEEEtran}
\bibliography{mybib}

\end{document}